\documentclass[runningheads]{llncs}
\usepackage[T1]{fontenc}
\usepackage{graphicx,verbatim}

\usepackage{amsmath}
\usepackage{booktabs} 
\usepackage{multirow} 
\usepackage{tabularx}
\usepackage{array}
\newcolumntype{C}{>{\centering\arraybackslash}X}
\usepackage{amssymb}

\usepackage{hyperref}
\usepackage{xcolor}
\usepackage{soul}
\begin{document}
\title{Uni-Light: An Ultra-Lightweight Framework via Uncertainty-Aware Knowledge Distillation for Brain Tumour Segmentation}
%

\author{
Libing Kuang\inst{1} \and
Soren Salehi\inst{2} \and
Ziling Wu\inst{1} \and
Ahmad P. Tafti\inst{3} \and
Armaghan Moemeni\inst{1}\thanks{Corresponding author}
}

\authorrunning{L. Kuang et al.}

\institute{
School of Computer Science, University of Nottingham,
Nottingham, UK
\and
Electrical Engineering Department, Sharif University of Technology,
Tehran, Iran
\and
University of Pittsburgh, Pittsburgh, PA, USA
}

\maketitle

\begin{abstract}
Accurate 3D brain tumour segmentation from multi-modal Magnetic Resonance Imaging (MRI) is essential for clinical diagnosis and treatment planning. Existing brain tumour segmentation methods often suffer from heavy computational demands, while current lightweight architectures frequently lack the capacity to maintain segmentation fidelity in complex tumour regions. To address these issues, we propose a novel ultra-lightweight framework (Uni-Light) that achieves high-fidelity segmentation with substantially reduced computational overhead. It combines multi-scale convolutions with an uncertainty-aware knowledge distillation scheme that directs the student model toward hard-to-classify regions, complemented by a Signed Distance Field boundary loss for geometric constraints. Experimental results on BraTS2023-GLI and MSD-BTS datasets demonstrate that Uni-Light reduces parameters by 97.56\%, floating-point operations (FLOPs) by 73.03\%, and inference memory footprint by 81.58\%, while surpassing the state-of-the-art model by an average of 1.47\% in Dice score, offering a highly competitive trade-off between segmentation accuracy and computational efficiency in resource-constrained clinical settings. This work also advances data engineering for medical imaging by demonstrating that teacher model uncertainty can be exploited as a data-driven supervisory signal, re-prioritising the training data distribution without requiring additional annotation. 

\keywords{Brain tumour \and Knowledge distillation \and Lightweight model.}

\end{abstract}

\section{Introduction}

Accurate 3D brain tumour segmentation from multi-modal MRI is essential for surgical planning and treatment monitoring. The BraTS benchmark~\cite{ref_brats} has substantially advanced automated tumour delineation, establishing deep neural networks as the dominant paradigm. Encoder--decoder convolutional architectures such as 3D U-Net~\cite{ref_3dunet} and its self-configuring extension nnU-Net~\cite{ref_nnunet} 
have become standard baselines for volumetric medical image segmentation.
While segmentation accuracy has steadily improved, this progress has largely been driven by increasing model capacity. Recent Transformer~\cite{ref_unetr} and Mamba-based~\cite{ref_mamba,ref_segmamba} architectures enhance contextual modelling but introduce prohibitive computational overhead for clinical deployment. To mitigate these overheads, lightweight designs employ efficient operators such as depthwise convolutions \cite{ref_mobilenet} or multi-scale feature fusion \cite{ref_dmfnet}. 

Knowledge Distillation (KD)~\cite{ref_kd,ref_kd_seg} provides a principled strategy for transferring knowledge from a high-capacity teacher to a compact student model, enabling compression without severe degradation in predictive performance. However, most existing distillation frameworks apply uniform supervision, overlooking the spatial heterogeneity of 3D brain tumours. By assuming homogeneous difficulty, these methods misallocate supervisory focus and degrade performance in ambiguous regions like tumor boundaries and enhancing components.

To address these limitations, we propose Uni-Light, an ultra-lightweight framework. With that, our main contributions are two-fold: \textbf{(1) Methodological:} We propose \textit{Uni-Light}, an ultra-lightweight framework for 3D brain tumour segmentation that combines a compact multi-scale architecture with uncertainty-aware knowledge distillation. By dynamically weighting supervision according to teacher uncertainty and incorporating boundary-aware optimisation, Uni-Light focuses learning on difficult tumour regions while maintaining extreme parameter efficiency. \textbf{(2) Clinical:} Through evaluation on the BraTS~\cite{ref_brats} and Medical Segmentation Decathlon (MSD)~\cite{ref_msd} benchmarks, we demonstrate that Uni-Light achieves competitive segmentation performance while substantially reducing model size, memory consumption, and computational cost. These results support the deployment of reliable brain tumour segmentation models in resource-constrained clinical environments and edge-computing settings.

The complete source code is publicly available for research, educational, and academic purposes only at  \url{https://github.com/libinglilykuang/UniLight-miccai}.

\section{Related Work}
Lightweight 3D segmentation networks are essential for clinical deployment, enabling real-time inference in resource-constrained environments without compromising accuracy. Efficient CNN-based backbones, such as LATUP-Net~\cite{ref_latup} and SuperLightNet~\cite{ref_superlightnet}, employ parallel convolutions and parameter aggregation to reduce redundant operations. Concurrently, lightweight Transformers like MAT~\cite{ref_mat} and N-LiNet~\cite{ref_nlinet} utilise axial or hybrid attention to mitigate the computational cost of long-range modeling. Emerging paradigms, notably DiffBTS~\cite{ref_diffbts}, further investigate the potential of lightweight diffusion models for 3D volumes. The above methods optimise 3D brain tumor segmentation by leveraging axial or hybrid attention mechanisms and parameter aggregation to mitigate computational overhead. In contrast, we explore the synergy between architectural efficiency and adaptive supervision via uncertainty-aware distillation.

KD was introduced to transfer softened probability distributions~\cite{ref_kd}, and has evolved to include feature-based hints~\cite{ref_fitnets}, attention transfer~\cite{ref_attention_transfer}, and structured alignment~\cite{ref_structured_kd}. In medical image segmentation, KD has been adopted to compress 3D convolutional networks and improve lightweight architectures. However, standard uniform distillation overlooks the spatial heterogeneity and predictive ambiguity in critical tumor boundaries and enhancing subregions, leading to suboptimal supervisory focus.
While common boundary-aware strategies, such as signed distance field losses \cite{ref_boundary_loss}, improve geometric precision beyond region-based overlap measures, their integration with spatially adaptive distillation remains underexplored. To bridge these gaps, we propose a novel lightweight framework designed to maintain high-fidelity segmentation performance even under strict computational constraints.

\section{Methodology: Uni-Light Framework}
\begin{figure}[!t]
    \centering
    \includegraphics[width=0.7\linewidth]{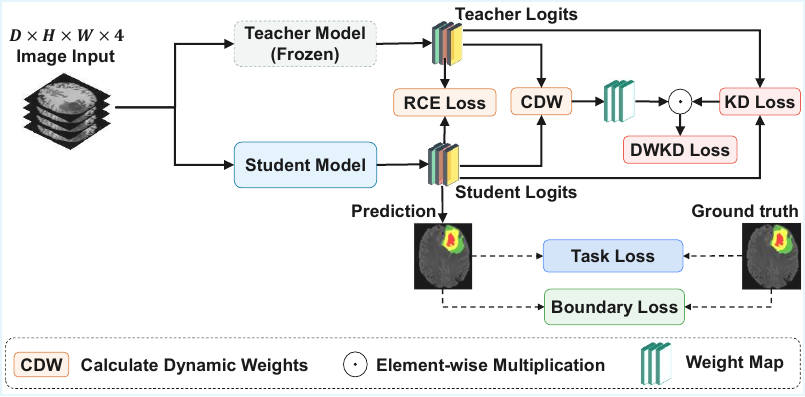}
    \caption{Overview of the proposed Uni-Light distillation framework. Its key components include Dynamic Weighted Knowledge Distillation (DWKD) and training losses: RCE loss, DWKD loss, task loss, and boundary loss.}
    \label{fig:architecture}
\end{figure}

Fig. \ref{fig:architecture} illustrates the Uni-Light framework, which comprises an efficient student architecture, an uncertainty-aware distillation mechanism, and a joint loss function for precise boundary delineation.

\subsection{Lightweight Student Architecture}
\begin{figure}[!t]
    \centering
    \includegraphics[width=0.9\linewidth]{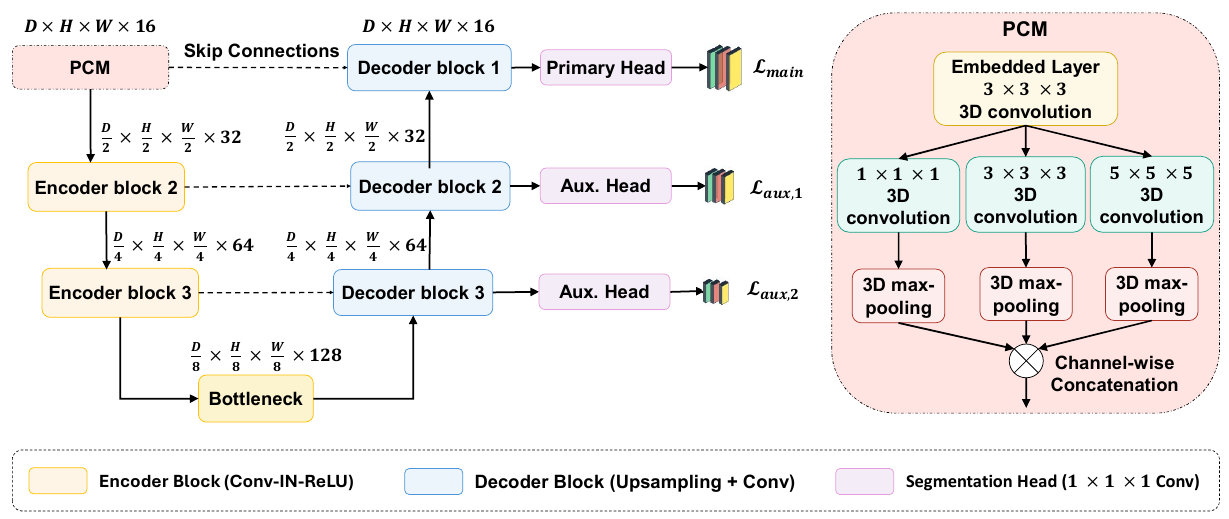}
    \caption{Detailed architecture of the student model. Its key components include the Parallel Convolution Module (PCM) for multi-scale feature extraction and a deep supervision mechanism with associated losses: $\mathcal{L}_{main}$, $\mathcal{L}_{aux,1}$, and $\mathcal{L}_{aux,2}$.}
    \label{fig:student}
\end{figure}

To achieve computationally efficient brain tumour segmentation, the Uni-Light framework is inspired by LATUP-Net \cite{ref_latup} and adopts its Parallel Convolutional Module (PCM) to construct a lightweight backbone network. This architecture is a 3D U-shaped structure that reduces the number of parameters by decreasing the network depth and feature channels, as illustrated in Fig.~\ref{fig:student}. To further optimise the balance between performance and efficiency, we improve the original architecture by removing the computationally intensive attention mechanism and introducing a deep supervision mechanism.

At the first layer of the backbone network, we integrate a parallel multi-scale convolution module. This module first uses a shared $3 \times 3 \times 3$ convolution layer to initially extract the input features, then branched into three parallel paths with kernel sizes $k \in \{1, 3, 5\}$ followed by 3D max-pooling and concatenation.

To address the potential gradient vanishing and facilitate faster convergence, we introduce a deep supervision mechanism \cite{ref_deep_supervision} within the decoder. In addition to the primary segmentation head at the highest resolution, two auxiliary heads are integrated into the intermediate stages of the decoder. These heads produce multi-scale predictions at $1/2$ and $1/4$ of the original resolution, which are then compared against the downsampled ground truth masks during training.

\subsection{Dynamic Weighted Knowledge Distillation}
To provide high-quality supervision, we employ a high-capacity Residual 3D U-Net as the teacher, following the nnU-Net~\cite{ref_nnunet} framework and trained on the MONAI~\cite{ref_monai} benchmark. The model comprises five resolution levels with downsampling strides of 2, utilising two residual units at each encoder and decoder stage to capture hierarchical features. The pre-trained teacher provides the softened probability maps and voxel-wise uncertainty estimates. Diverging from static distillation, Dynamic Weighted Knowledge Distillation (DWKD) adopts voxel-wise weights by quantifying the teacher's predictive uncertainty.

Due to the heterogeneity of the lesions and the ambiguity of boundaries in brain tumour segmentation, the prediction confidence of the teacher model varies significantly for different regions. We use information entropy to quantify the uncertainty. Let $P_{t,i}$ be the softened distribution of teacher logits $t_i$ with temperature $\tau$ \cite{ref_kd}; the voxel-level uncertainty $H_i$ is derived as:
\begin{equation}
P_{t,i} = \sigma\left(\frac{t_{i}}{\tau}\right), \quad H_{i} = - [ P_{t,i} \ln(P_{t,i}) + (1 - P_{t,i}) \ln(1 - P_{t,i}) ]
\end{equation}
where $\sigma(\cdot)$ denotes the Sigmoid function. High $H_i$ values typically localise to tumour boundaries or complex tissues where the teacher is less certain.

Based on the calculated uncertainty $H_i$, we assign dynamic weights $\alpha_i$ to each voxel in the loss function. For training stability, we normalise $H_i$ by the maximum entropy in the current batch:
\begin{equation}
\alpha_i = \frac{H_i}{\max(\{H_1, H_2, \dots, H_L\}) + \epsilon}
\end{equation}
where $L$ represents the total number of voxels in the current batch and $\epsilon$ is a very small constant used to prevent the denominator from being zero.

\subsection{Loss Function}
To ensure precise segmentation and effective knowledge transfer, the Uni-Light framework is optimised via a composite objective function:
\begin{equation} 
\mathcal{L}_{\text{total}} = \lambda_1 \mathcal{L}_{\text{task}} + \lambda_2 \mathcal{L}_{\text{DWKD}} + \lambda_3 \mathcal{L}_{\text{RCE}} + \lambda_4 \mathcal{L}_{\text{bound}} 
\end{equation} 
where the coefficients are empirically set to $\lambda_1=0.6, \lambda_2=0.2, \lambda_3=0.1$, and $\lambda_4=0.1$ to balance the gradient magnitudes from different supervision sources. To handle class imbalance and multi-scale convergence, we formulate the task loss in a single objective:
\begin{equation}
\mathcal{L}_{\text{task}} = \sum_{l=1}^{3} \eta_l \sum_{c \in \{WT, TC, ET\}} v_c (\mathcal{L}_{\text{BCE}, c}^{(l)} + \mathcal{L}_{\text{Dice}, c}^{(l)})
\end{equation}
where $c \in \{WT, TC, ET\}$ and $l \in \{1, 2, 3\}$ represent tumour regions and deep supervision levels, respectively. The hierarchical weights are set to $\eta = \{1, 0.5, 0.25\}$, and region-specific penalties are $v = \{1, 1, 3\}$, prioritising the challenging enhancing tumour region.

We employ DWKD to distill semantic knowledge and RCE \cite{ref_dwkd} to suppress background noise:
\begin{equation}
\mathcal{L}_{\text{DWKD}} = \frac{1}{N}\sum_{i=1}^{N} \alpha_i \cdot \text{KL}(P_{t,i} \parallel P_{s,i}), \quad
\mathcal{L}_{\text{RCE}} = - \frac{1}{N} \sum_{i=1}^{N} A_i \log(P_{s,i})
\end{equation}
where $\alpha_i$ is the voxel-wise dynamic weight from Eq.~(2), $\text{KL}(\cdot \parallel \cdot)$ denotes the Kullback-Leibler divergence, and $A_i \in \{0,1\}$ is a confidence-thresholded mask derived from the teacher's predictions that determines whether voxel $i$ participates in the RCE loss ($A_i=1$) or is ignored ($A_i=0$), effectively suppressing confident background regions while focusing learning on uncertain areas.

To preserve edge integrity, we introduce a boundary loss based on the SDF~\cite{ref_boundary_loss}. For each ground-truth mask, we precompute a Signed Distance Map $S_{GT}$ where each voxel $i$ is assigned $S_{GT}(i) = -d(i,\partial\Omega)$ inside the tumour and $+d(i,\partial\Omega)$ outside, with $d(i,\partial\Omega)$ denoting the Euclidean distance to the nearest boundary point $\partial\Omega$. The boundary loss is then formulated as:
\begin{equation}
\mathcal{L}_{bound} = \frac{1}{N}\sum_{i=1}^{N} p_i \cdot S_{GT}(i)
\end{equation}
This penalises false positives in the background ($S_{GT}(i) > 0, p_i \approx 1$) and rewards correct predictions inside the tumour ($S_{GT}(i) < 0, p_i \approx 1$), providing continuous geometric supervision that guides the model to refine tumour boundaries.

\section{Experiments}
\subsubsection{Datasets.} We evaluate on BraTS2023-GLI~\cite{ref_brats23} (1,251 cases) and MSD BTS~\cite{ref_msd} (484 cases). Each case provides four MRI sequences (T1, T1ce, T2, FLAIR) with expert annotations for three tumour sub-regions: necrotic core (NCR), peritumoural edema (ED), and enhancing tumour (ET). Following the standard protocol, we report results on ET, tumour core (TC), and whole tumour (WT).

\subsubsection{Evaluation Metrics.} 
We evaluate both segmentation accuracy and computational efficiency. Accuracy is measured by Dice Similarity Coefficient (DSC)~\cite{ref_dice} and 95\% Hausdorff Distance (HD95)~\cite{ref_hd95}, where HD95 evaluates the model's precision in boundary delineation by calculating the 95th percentile of the distances between the surface points of the predicted and ground truth masks. For lightweight assessment, we report: 1) Parameters (Pars.), the memory footprint of model weights; 2) FLOPs, the computational cost per inference; 3) Inference Latency (Lat.), the average time per scan; and 4) inference memory (IM) and training memory (TM), the peak memory usage during deployment and training, respectively.

\subsubsection{Implementation Details.} 
We use an AdamW optimiser \cite{ref_adamw} with an initial learning rate of $1 \times 10^{-4}$ and a weight decay of $1 \times 10^{-5}$ to ensure stable convergence during knowledge transfer. Training runs on a single NVIDIA T4 GPU (batch size 2) with five-fold cross-validation. All volumes are resampled to $128^3$ voxels, with on-the-fly augmentation including random flipping, $90^{\circ}$ rotations, intensity scaling, and Gaussian noise.

\subsubsection{Comparison with SOTA Methods.}
\begin{figure}[!t]
    \centering
    \includegraphics[width=0.8\textwidth]{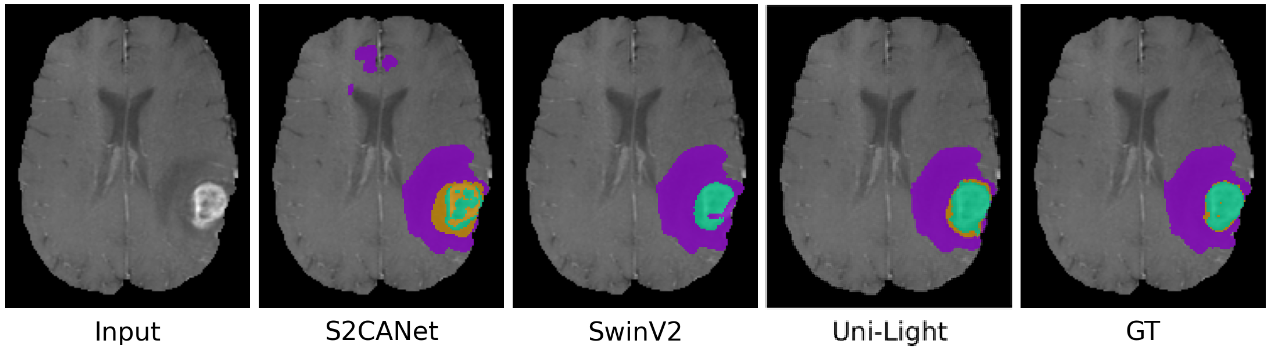}
    \caption{Visual comparisons of Uni-Light and several leading baselines on the MSD-BTS dataset. The brown, purple, and green regions denote peritumoural edema (ED), necrotic tumour core (NCR) and enhancing tumour (ET), respectively.}
    \label{fig:visualization}
\end{figure}

As shown in Table~\ref{tab:evaluation_merged}, Uni-Light achieves SOTA performance with average DSC scores of 89.97\% on BraTS2023-GLI and 84.40\% on MSD-BTS, notably surpassing BraTS-UMamba in challenging ET and TC sub-regions. While Uni-Light exhibits a higher HD95, this is an expected trade-off for its extreme 37$\times$ parameter reduction compared to heavy models. Table~\ref{tab:complexity} highlights our model's efficiency, outperforming SwinUNETR and SegMamba in latency by 3.5$\times$ and 4.7$\times$, respectively. With a minimal 4059 MB training footprint, Uni-Light avoids the OOM issues typical of heavy architectures on standard hardware. In Figure~\ref{fig:visualization}, we also visually compare Uni-Light and several leading baselines, demonstrating that our method achieves superior segmentation fidelity and sharper boundaries.

\begin{table}[!t]
\centering
\caption{Quantitative evaluation on BraTS2023 and MSD-BTS datasets. Results for baseline methods are cited from \cite{ref_brats_umamba}. Best results are in \textbf{bold}.}
\label{tab:evaluation_merged}
\begin{tabularx}{\textwidth}{l*{8}{C}}
\toprule
\multirow{2}{*}{Methods} & \multicolumn{4}{c}{DSC (\%) $\uparrow$} & \multicolumn{4}{c}{HD95 (mm) $\downarrow$} \\
\cmidrule(lr){2-5} \cmidrule(lr){6-9}
 & ET & WT & TC & Avg. & ET & WT & TC & Avg. \\
\midrule
\multicolumn{9}{c}{\textit{BraTS2023-GLI Dataset}} \\
\midrule
Res-UNet (Teacher) \cite{ref_monai,ref_resunet_base}& 76.08 & \textbf{92.92}  & 87.98& 85.66 &3.53 & 4.38 &  4.00 & 3.97 \\
Eoformer \cite{ref_eoformer} & 83.11 & 90.68 & 87.60 & 87.13 & 4.60 & 7.50 & 5.92 & 6.01 \\
SDV-TUNet \cite{ref_sdv} & 83.96 & 90.50 & 87.36 & 87.27 & 3.41 & 5.61 & 5.29 & 4.77 \\
S$^2$CA-Net \cite{ref_s2ca} & 83.91 & 91.60 & 87.91 & 87.81 & 4.46 & 6.94 & 6.04 & 5.82 \\
SwinUNETR-V2 \cite{ref_swinv2} & 84.07 & 91.85 & 88.12 & 88.01 & 3.92 & 5.86 & 5.33 & 5.04 \\
SegMamba \cite{ref_segmamba} & 84.65 & 92.12 & 88.34 & 88.37 & 4.22 & 4.92 & 5.80 & 4.98 \\
BraTS-UMamba \cite{ref_brats_umamba} & 85.74 & 92.90 & 90.71 & 89.78 & \textbf{3.11} & \textbf{4.09} & \textbf{3.80} & \textbf{3.66} \\
Our Uni-Light & \textbf{86.59} & 92.59 & \textbf{90.74} & \textbf{89.97} & 7.23 & 4.78 & 4.25 & 5.42 \\
\midrule
\multicolumn{9}{c}{\textit{MSD-BTS Dataset}} \\
\midrule
Res-UNet (Teacher) \cite{ref_monai,ref_resunet_base}& 71.08 & 87.92  & 82.98& 80.66 &3.82 & 5.50 &  5.12 & 4.82 \\
Eoformer \cite{ref_eoformer} & 74.28 & 88.29 & 80.95 & 81.17 & 5.98 & 8.59 & 7.10 & 7.23 \\
SDV-TUNet \cite{ref_sdv} & 73.42 & 87.69 & 79.59 & 80.23 & 5.96 & 7.09 & 7.52 & 6.86 \\
S$^2$CA-Net \cite{ref_s2ca} & 77.35 & 89.40 & 82.68 & 83.14 & 5.62 & 7.49 & 7.54 & 6.88 \\
SwinUNETR-V2 \cite{ref_swinv2} & 75.92 & 88.69 & 82.28 & 82.50 & 5.45 & 8.03 & 6.96 & 6.81 \\
SegMamba \cite{ref_segmamba} & 76.82 & 89.62 & 82.74 & 83.06 & 5.31 & 7.37 & 6.45 & 6.38 \\
BraTS-UMamba \cite{ref_brats_umamba} & \textbf{80.63} & \textbf{90.61} & \textbf{84.08} & \textbf{85.11} & \textbf{3.92} & \textbf{4.93} & \textbf{5.14} & \textbf{4.66} \\
Our Uni-Light & 79.00 & 90.50 & 83.67 & 84.40 & 9.30 & 6.14 & 6.07 & 7.17 \\
\bottomrule
\end{tabularx}
\end{table}

\begin{table}[!t]
\small
\setlength{\tabcolsep}{3pt}
\centering
\caption{Computational complexity analysis with $4 \times 128^3$ input resolution. ``OOM'' indicates out of memory. Best results are in \textbf{bold}.}
\label{tab:complexity}
\begin{tabular}{lccccc}
\toprule
Model & Pars. (M) & FLOPs (G) & Lat. (ms) & IM (MB) & TM (MB) \\ 
\midrule
Res-UNet (Teacher) \cite{ref_monai,ref_resunet_base} & 19.22 & \textbf{105.82} & 423.88 & \textbf{622} & \textbf{1820} \\
Eoformer \cite{ref_eoformer} & 5.13 & 133.67 & 320 & 2118 & 6002 \\
SDV-TUNet \cite{ref_sdv} & 20.80 & 252.95 & 397 & 5616 & 11513 \\
S$^2$CA-Net \cite{ref_s2ca} & 25.18 & 525.58 & \textbf{274} & 2374 & 4265 \\
SwinUNETR-V2 \cite{ref_swinv2} & 62.19 & 793.92 & 1017 & 8400 & OOM \\
SegMamba \cite{ref_segmamba} & 67.42 & 2920.00 & 1368 & 5575 & OOM \\
BraTS-UMamba \cite{ref_brats_umamba} & 56.89 & 323.92 & 517 & 2057 & 3874 \\ 
Our Uni-Light & \textbf{1.52} & 214.10 & 289 & 1547 & 4059 \\ 
\bottomrule
\end{tabular}
\end{table}

\begin{table}[!t]
\centering
\caption{Ablation study of investigating key components in our model on the BraTS2023 dataset. The best result is indicated in \textbf{boldface}.}
\label{tab:ablation_study}
\begin{tabularx}{\textwidth}{l*{5}{C}}
\hline
 & baseline & S1    & S2    & S3    & ours\\ 
 \hline
DSC (\%) $\uparrow$    & 86.38    & 87.08 & 88.05 & 89.31 & \textbf{89.95} \\
HD95 (mm)~$\downarrow$ & 5.52     & 5.14  & 4.23  & \textbf{3.88}  & 6.29           \\ \hline
\end{tabularx}
\end{table}

\subsubsection{Ablation Study.}Table~\ref{tab:ablation_study} evaluates the contribution of each core component in Uni-Light. We adopt a 3D U-Net student model trained with KD as our baseline. The integration of DWKD (S1) improves regional segmentation fidelity by modulating supervision according to the teacher's predictive uncertainty. Adding geometric boundary constraints through the SDF-based loss (S2) further refines the tumour contours. The introduction of the PCM (S3) demonstrates substantial performance enhancement by fusing multi-scale context information. 

Finally, incorporating RCE (Ours) achieves the optimal DSC (89.95\%) but yields a higher HD95 (6.29 mm) compared to S3 (3.88 mm). We attribute this to RCE suppressing background noise to improve global mask coverage, which can introduce sparse boundary outliers that disproportionately inflate the HD95. This is consistent with the original DWKD framework~\cite{ref_dwkd}. Crucially, the boundary loss (S1→S2) reduces HD95 from 5.14 to 4.23 mm, confirming its geometric regularisation effect; the final HD95 increase is thus RCE-specific and orthogonal to boundary precision. Given that DSC is the primary metric for clinical tumour volumetry and that the observed HD95 gap falls within the range of inter-rater variability in manual annotation, we regard this trade-off as clinically acceptable.

\section{Conclusion and Discussion}

We presented Uni-Light, an ultra-lightweight framework for 3D brain tumour segmentation that combines multi-scale convolutions, uncertainty-aware knowledge distillation, and boundary-aware optimisation. Experiments on BraTS2023 and MSD-BTS show that Uni-Light maintains high segmentation fidelity while substantially reducing parameters and computational cost. While the teacher model influences the student's performance ceiling, Uni-Light's inference efficiency depends solely on the student architecture, ensuring consistent computational advantages in clinical deployment. Beyond its efficiency, Uni-Light demonstrates that teacher uncertainty can serve as a data-driven signal to re-prioritise training samples without additional annotation.

While Uni-Light achieves a favourable accuracy–efficiency trade-off, its reliance on a single teacher model currently constrains the quality of uncertainty estimation, and the emphasis on volumetric accuracy introduced by the RCE loss can lead to increased boundary errors in challenging cases. These boundary errors may be mitigated through complementary strategies such as post-processing. Future work will also focus on stabilising supervision via stronger teacher architectures with attention mechanisms, and validating the generalisability of Uni-Light across additional medical imaging tasks.

\begin{credits}

\subsubsection{\ackname}
The authors gratefully acknowledge the academic and institutional support provided by the University of Nottingham and the University of Pittsburgh. The authors would like to acknowledge Dr Stefanie Thust, School of Medicine, University of Nottingham, for her valuable clinical advice. We also thank Denilson Tafa, PhD student in the School of Computer Science, University of Nottingham, for his critical review of the literature review. The authors also thank the organisers and contributors of the BraTS 2023
Challenge and the Medical Segmentation Decathlon for making the datasets
used in this study available.

\subsubsection{\discintname}
The authors declare that they have no competing interests relevant to the content of this article.

\end{credits}

\end{document}